# Curve-based Neural Style Transfer

**Yu-hsuan Chen, Levent Burak Kara, Jonathan Cagan**
Carnegie Mellon University
{yuhsuan2, lkara, jcag}@andrew.cmu.edu

## Abstract

This research presents a new parametric style transfer framework specifically designed for curve-based design sketches. In this research, traditional challenges faced by neural style transfer methods in handling binary sketch transformations are effectively addressed through the utilization of parametric shape-editing rules, efficient curve-to-pixel conversion techniques, and the fine-tuning of VGG19 on ImageNet-Sketch, enhancing its role as a feature pyramid network for precise style extraction. By harmonizing intuitive curve-based imagery with rule-based editing, this study holds the potential to significantly enhance design articulation and elevate the practice of style transfer within the realm of product design.

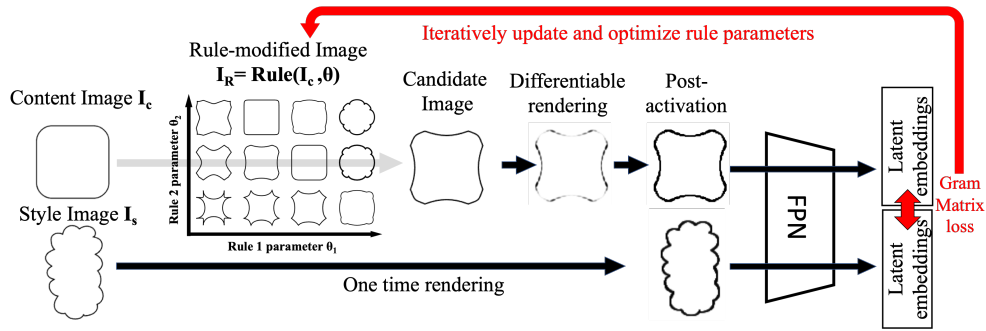

Figure 1: Workflow of the proposed curve-based style transfer method.

## 1 Introduction

The process of designing product exteriors often begins with sketch-like, curve-based images [1–6], where aesthetic and stylistic considerations frequently emerge as explicit rule-based operations [7–10]. However, conventional neural style transfer methods [11], while successful in converting textures from a given style image to the content image, often struggle to significantly alter the shapes of binary sketches according to a given style image. Bridging this gap, this research introduces a novel parametric rule-based style transfer framework that operates directly on scalar vector graphics (SVGs), with a primary focus on the style transformation of curves. By combining the intuitive nature of curve-based imagery with the explicit curve editing rules, this approach offers a promising avenue for enhancing style transfer in design, aligning more closely with how designers conceptualize and articulate their ideas through curves.

## 2 Method

The goal of this research is to perform style transfer directly on SVGs in a rule-based workflow, shown in Figure 1. The key components and challenges are elaborated in the following paragraphs.

**Rules Design for Effective Shape Editing.** The initial challenge centers on creating curve-editing rules that are both prominent yet stable. This research addresses this challenge by devising differentiable modification rules encompassing rigid body motions, shear, curvature alterations, and smoothing operations, all contributing to the attainment of the desired outcomes.

**Sketch to Pixelated Canvas Transformation.** The second challenge involves transforming curves into pixelated canvases in a differentiable and efficient manner without overburdening the GPU. To tackle this, differentiable rendering is adopted and modified to rasterize curves using ReLU activation function [12]. In practice, SVG data are homogenized into cubic Bezier representations, and for each Bezier curve, N+1 control points are sampled to approximate the curve with N line segments. Given the impracticality of rendering curves one at a time using the CPU, batch rendering is opted. To further mitigate potential GPU overload for generic SVGs that can have numerous curves, dropout is also used during the optimization process.

**Fine-Tuning Feature Pyramid Network (FPN).** After rendering both style and content images, an FPN is used to calculate style similarity. It is found that pretrained VGG19 [13] is not the ideal FPN for sketch style extraction. To improve the outcomes, this research fine-tunes VGG19 on ImageNet-Sketch data [14], as it most closely aligns with binary sketch images.

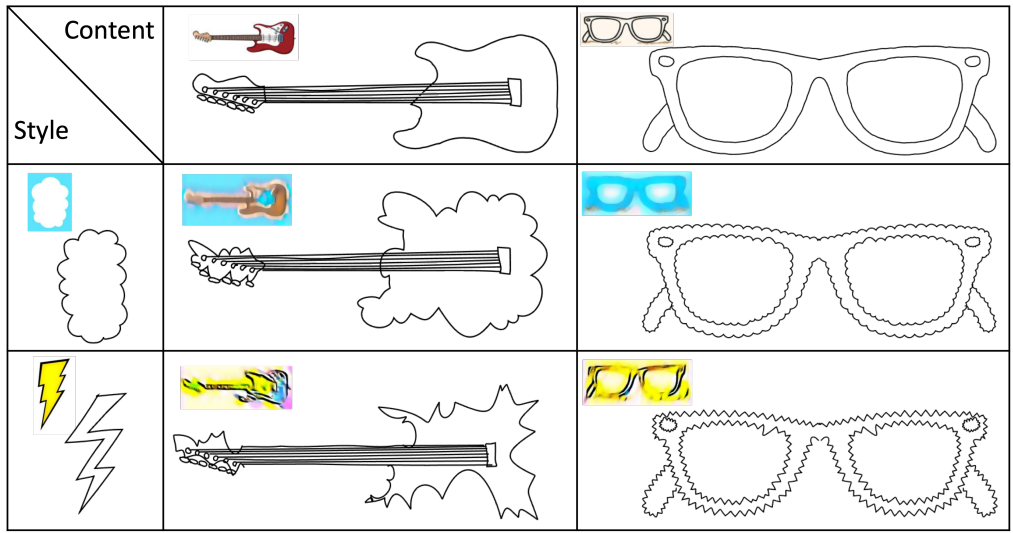

Figure 2: Transferring styles of cloud and lightning to the content of guitar and glasses, using shape rules including control point translation and curvature modification. Our style-transferred results are shown as binary sketches, and the fully pixel-based neural style transfer result are shown at the upper left corners.

## 3 Results and Conclusions

Figure 2 shows and compares style transfer results between curve-based and pixel-based methods (Image sources: [15–17]). Notably, our approach excels in translating style objectives into content images, demonstrating a superior ability to modify product shapes. In contrast, pixel-based neural style transfer primarily focuses on transferring color textures without substantial shape alteration.

Conclusively, this research showcases the potential of our innovative style transfer framework for SVGs, surpassing traditional neural style transfer. By directly manipulating SVGs using parametric curve-editing rules, differentiable rendering curves as multiple line segments, and a fine-tuned VGG19 FPN on sketch images, this approach aims to automatically convey and transfer SVGs' styles by emphasizing curve manipulation.

This work bridges the realms of design aesthetics and computational artistry, offering a promising avenue for curve-based imagery and design in style transfer. Future directions involve exploring advanced shape rules, enhancing stability, imposing geometric constraints, venturing into 3D applications, and conducting human perceptual studies to further enrich design aesthetics.

## 4 Ethical Implications

Because of the transformative capabilities this methodology offers in design articulation and style transfer within the realm of product design, it may carry important ethical implications. First and foremost is the concern regarding the displacement of designers' roles. It is crucial to clarify that this technology is not designed to replace designers but rather to augment their capabilities. Design experts remain essential in making intelligent decisions, selecting appropriate design rules, and creatively using this generative design tool to explore novel ideas.

Another ethical dimension pertains to copyright and intellectual property. It is imperative that users adhere to ethical standards and cite both the original source images and the style images when sharing the generated content. This practice not only safeguards the rights of creators but also ensures responsible and ethical use of the technology, addressing potential copyright issues effectively.